\pdfoutput=1

\documentclass[11pt]{article}

\usepackage[]{emnlp2021}

\usepackage{times}
\usepackage{latexsym}
\usepackage{hhline}
\usepackage{graphicx}
\usepackage{amsmath,amsfonts}
\usepackage{float}
\usepackage{caption}
\usepackage{subcaption}

\usepackage[T1]{fontenc}

\usepackage[utf8]{inputenc}

\usepackage{microtype}

\title{Monitoring geometrical properties of word embeddings for detecting the emergence of new topics}

\author{Clément Christophe \\
  Université de Lyon, Lyon 2\\
  UR ERIC, France \\
  EDF R\&D, Palaiseau, France \\
  \texttt{clement.christophe@edf.fr}\\\And
  Julien Velcin \and Jairo Cugliari \\
  Université de Lyon, Lyon 2\\
  UR ERIC, France \\
  \texttt{julien.velcin@univ-lyon2.fr}\\
  \texttt{jairo.cugliari@univ-lyon2.fr}\\\AND
    Manel Boumghar \and Philippe Suignard \\
    EDF R\&D, Palaiseau, France\\
    \texttt{manel.boumghar@edf.fr}\\
    \texttt{philippe.suignard@edf.fr}

  } 

\date{}

\begin{document}
\maketitle
\begin{abstract}
Slow emerging topic detection is a task between event detection, where we aggregate behaviors of different words on short period of time, and language evolution, where we monitor their long term evolution. In this work, we tackle the problem of early detection of slowly emerging new topics. To this end, we gather evidence of weak signals at the word level. We propose to monitor the behavior of words representation in an embedding space and use one of its geometrical properties to characterize the emergence of topics. As evaluation is typically hard for this kind of task, we present a framework for quantitative evaluation.
We show positive results that outperform state-of-the-art methods on two public datasets of press and scientific articles.
\end{abstract}

\section{Introduction}

For a company receiving hundreds of thousands of client feedbacks per month, it is crucial to analyze the dynamics of the textual content in an efficient way. While it is common to detect events and sudden bursts, slowly emerging new topics are often detected too late. Early detection of new topics could lead a company to better understand their clients' feedback, to detect implicit problems in their infrastructure that can cause problems for certain types of clients and then anticipate marketing or communication responses.

We consider that detecting emerging topics is a task close to the fields of \textit{Event detection} and \textit{Linguistic change detection}. The former focuses on detecting groups of words (e.g., topics) that are evolving fast in the form of a burst. The latter focuses on analyzing single word meanings evolving slowly. In this work, we are interested in detecting lexical fields that are evolving slowly over time (i.e., topics becoming dominant). 

In this work, we represent our textual data in the form of words in an embedding space and observe their evolution through time. We notice that single words evolve differently in high dimensional space according to the type of temporal dynamic they are linked to. Instead of quantifying change of meaning and polysemy, we analyze changes through the scope of events and slow emergence of topics. During our observations, we notice that a positive correlation is linked with event-like topics while a negative correlation is a sign of emergence (see Figure~\ref{fig:correlation}). We develop a method to detect words associated with these topics as early as possible. Additionally, we develop a framework to artificially introduce emerging topics into our data in order to build a gold standard for detection. 

After reviewing the literature on the subject (Section \ref{sec:related}), we present a framework to evaluate our task by simulating the dynamic of novelty in a textual dataset (Section \ref{sec:novelty}). We develop a system based on our intuition that there is a specific correlation between word frequencies and movement in an embedding space (Section \ref{sec:method}). Then, we explain how we got this intuition and illustrate it with two datasets (Section \ref{sec:experiments}). Finally, we show that our method is the best in terms of qualitative and quantitative results with respect to state-of-the-art baselines of the literature (Section \ref{sec:results}).

\section{Related Works}\label{sec:related}

Events are discussed in news stories every day as they describe important change in the world. 
\textit{Event Detection} in news stories has long been addressed in the Topic Detection and Tracking (TDT) program \cite{allan1998topic}. Several approaches have been studied to solve this task \cite{allan2000first, sayyadi2009event,kumaran2004text,petrovic2010streaming,lau2012line,cordeiro2012twitter, nguyen2016joint, liu2017exploiting, ghosal-etal-2018-novelty}.

On a longer time frame, textual data has also been studied through the scope of language evolution. Change of meaning in words are the consequences of the appearance of new concepts, of the emergence of new technologies that change our way of life and then, our way of talking. Topic evolution, which represents how words are mixed together, has been studied with the help of topic modeling algorithms like Latent Dirichlet Allocation (LDA) \cite{blei2006dynamic} and its extension in \cite{blei2006dynamic,wang2012continuous,wang2006topics}. Recently, the emergence of embedding representations, that represent word meanings, in textual data like Word2Vec \cite{mikolov2013distributed} has boosted the field of studies around language evolution. Works like \cite{kim-etal-2014-temporal,kulkarni2015statistically,hamilton2016diachronic,dubossarsky2017outta} have analyzed the structure of vector representation of words in order to illustrate their change of meaning through a long period of time. In \cite{hamilton2016diachronic} and \cite{dubossarsky2017outta}, they quantify the correlation between movement in an embedding space and frequency of a word by viewing it through the scope of semantic change and polysemy. Finally, contextual language models such as BERT \cite{devlin2018bert} and ELMO \cite{peters2018deep} have considerably improved the field of NLP. While it is known that this type of models slightly improve results in the task of semantic changes, they are also very time and resources consuming when we are monitoring an entire vocabulary \cite{kutuzov-giulianelli-2020-uio, Martinc2020LeveragingCE}.

In the literature, some works as \cite{huang2015topic}, \cite{peng2018emerging} and \cite{saeed2019enhanced} focus on detecting emerging topics with applications on Twitter. A work like \cite{asooja2016forecasting} try to predict the future distribution of words using tf-idf scores in order to predict emerging topics. Events are characterized by high and sudden bursts in the data. Their appearance is easily detectable on a micro time scale. Linguistic changes are slow, taking several years to change and can be illustrated on a macro time scale. In this work, we focus on the task of detecting as soon as possible weak signals in word movements associated with the slow emergence of a new topic.

\section{Novelty Definition}\label{sec:novelty}

The underlying task to resolve in the field of Novelty Detection is ill-posed. There is no clear consensus on the definition of ``novelty'' and no existence of a general framework for evaluation \cite{amplayo-etal-2019-evaluating}. In common sense, the term novelty refers to something that has not been observed before, therefore a single point in the data. In the literature, novelty is often linked to a signal \cite{eckhoff2014detecting} and corresponds to an unexpected evolution of it. The work of \cite{pimentel2014review} presents a review of novelty detection in various fields and makes the distinction between Novelty Detection and Anomaly or Outlier Detection. Anomaly and outlier are single-point observations in a dataset while novelty (especially emerging novelty) corresponds to a series of small anomalies that leads to a brand new cluster or topic. In this work we define novelty as follows: novelty is observed when an underlying significative change in the distribution of the data is detected. Textual data can be modeled as a word, topic or document. We develop a method for detecting, as early as possible, a slow emerging new topic by observing changes in word meaning. At the beginning of our observation, this topic is almost non-existent, that is, mixed up with noise. Over time, this topic grows slowly and finally become a major topic in our data.

Evaluation for novelty detection is challenging: no annotated dataset for this task exists in the literature and quantitative evaluation is essential to support the domain. With that idea in mind, we propose to artificially insert novelty in a dataset in order to simulate the emergence of a new topic. Datasets with annotated categories associated with each document are common in the NLP literature. We select one category in our dataset (for example: category about Basketball or Theater in the \textit{New York Times}) and re-order each of its documents with regards to time. This category now acts as a (controlled) emerging topic. The rate at which we introduce new documents into the dataset is defined as a logistic function, which is given by:
\begin{equation}
    r(t) = \frac{K}{1+\alpha e^{-rt}}, \qquad t,\alpha \in \mathbb{R}, 
\end{equation}
where $K>0$ and $r > 0$. Parameter $\alpha$ controls when the signal reaches 50\% of its final volume. We set $\alpha=1$ in order to have a centered emergent signal. The function is monotonically increasing from 0 to $K$. It allows to describe novelty as a rapid growth at geometric rates during the first introduction. At the saturation stage, the growth slows down to arithmetic rates until maturity. Parameter $r$ allows us to control the speed of the emergence. In this work, we experiment with $r = 0.3$ for slow emergence, $r = 0.5$ for normal emergence and $r = 1.0$ for fast emergence.

With one category dynamics matching exactly this signal, we make sure to have a quantifiable gold standard as a proxy of ideal ground truth. This approach allows us to organize which category will act as our emerging novelty, to quantify the importance of the rate at which the novelty emerges. There is only one category introduced as novelty during each experience but we are repeating our experiences several times on several categories in order to have stable results.

\section{CEND Methodology}\label{sec:method}

The use of embedding representation spaces for analyzing language evolution is now a common field in the NLP literature. In this kind of representation, similar entities (e.g., words or topics) are close if they are used in the same context: their meaning depends on the entities around them. Over time, their meaning may change, therefore their representation in the embedding space may be modified: there is a movement. In this work, we consider one type of movement in this space.

Let $\mathcal{W} = \lbrace w_1, \ldots, w_n \rbrace$ be a set of words. For each word 
$w \in \mathcal{W} $ 
we look at its numerical representation by a dense vector, say $v^w\in \mathbb{R}^D$, using a word 
embedding algorithm, say $\mathcal{A}$ (e.g., \texttt{SVD}, \texttt{word2vec}, \texttt{Glove}, \texttt{fasttext}). 
Typically, we examine how
the vector $v^w$ changes when consecutive bunches of documents are used to update the 
parameters of model $\mathcal{A}$.

Let us consider $v_1^w, v_2^w, \ldots, v_T^w$ a sequence of vectors representing the word
$w$ at each time slice $t = 1, \ldots, T$. Now, we define a measure to quantify the changes in the representation of the word $w$.

\paragraph{Magnitude of the change.} For two consecutive vectors, we look at the size of the
change using the euclidean distance:
\begin{equation}\label{eq:movement}
d_w (v_t^w, v_{t-1}^w) = \|v_t^w - v_{t-1}^w\|.
\end{equation}
In Section \ref{sec:experiments}, we show that a negative correlation seems to exist between words movement in an embedding space and its frequency \emph{if this word is part of an emerging topic}. Following this hypothesis, we based our algorithm on this correlation and we monitor its evolution through time for each word in our vocabulary. As our goal is to detect as soon as possible if a word is a part of an emergent topic, we monitor the Spearman correlation $\rho_{X,Y}$ between movement $d_w$ and frequency $f_w$ on a portion of the signal:
\begin{equation}\label{eq:correlation}
    \rho_{rf_w,rd_w}^t = \frac{\hbox{cov}(rf_w^{[t-n,t]},rd_w^{[t-n,t]})}{\sigma_{rf_w^{[t-n,t]}}\sigma_{rd_w^{[t-n,t]}}},
\end{equation}
where $f_w^t$ is the frequency of word $w$ at time $t$ and $rf_w$ and $rd_w$ denote the rank variables of $f_w$ and $d_w$ series. $n$ is the size of the sliding window. $cov$ corresponds to the covariance and $\sigma$ to the standard deviation. A word $w$ is considered emergent at time $t$ when its $\rho_w^t < k$, where $k$ is a threshold defined in section \ref{sec:threshold}. We call this method Correlation-based Embedding Novelty Detection (CEND).

\section{Experimental Setup}\label{sec:experiments}

In this work, we focus on the correlation between word frequency and word movement in the chosen embedding space. By artificially inserting documents related to an emerging new topic in a corpus as described in Section \ref{sec:novelty}, we notice that the amplitude of the movement of a word in the embedding space is linked with the dynamic of its frequency. In this section, we explain how correlation between these two measures and the dynamic of a topic are linked in a textual corpus.

\begin{table*}
\centering
    \begin{tabular}{cccccc}
        \hline
       dataset & docs & language & \# of used cat. & time range & size of vocabulary \\
       \hline
       NYTAC & 300K & English & 13 & 1995-2005 & 20.000\\
       SCI & 8337 & English & 4 & 1990-2005 & 5.000\\
       \hline
    \end{tabular}
    \caption{Summary of datasets used for this work.}
    \label{tab:datasets}
\end{table*}

\subsection{Data}

We work with 2 different datasets: the \textit{New York Times Annotated Corpus} \footnote{\url{https://catalog.ldc.upenn.edu/LDC2008T19}} (NYTAC) and a corpus of scientific abstract from AMiner\footnote{ \url{https://www.aminer.org/data}} (SCI). Documents in these corpora are associated, manually or automatically, with categories and we use these categories, as presented in Section \ref{sec:novelty}, to simulate the emergence of a new topic. A summary of these datasets is presented in Table \ref{tab:datasets}. 

The time-step size corresponds to a month for NYTAC and a year for SCI. Words are lower-cased and lemmatized. Punctuation and numerals are removed from the data.

\subsection{Building a Gold-Standard}

In order to evaluate quantitatively our approach CEND for detecting emerging topics, we need to construct a gold standard: words that we want to detect. As we said in Section \ref{sec:novelty}, we artificially introduced some categories through time and we want to detect weak signals, in the form of words, that are carried by the categories. Independently of our simulation, we train a Naive Bayes classifier on the entire dataset and we extract the 100 most discriminative features (i.e,  words) for each of the categories. 100 words have been selected in order to obtain statistically significant results for each category while limiting the number of non-specific words into the Gold-Standard. Results of the classification are not detailed in this work but as the accuracy comes close to 80\% on each dataset, we argue that it is enough to extract meaningful words. Some of these words are illustrated in Table~\ref{tab:discriminative}: they correspond to the words we want to automatically detect for each introduced category.

\begin{table*}
\centering
    \begin{tabular}{c|c|c|c|c|c}
        \hline
       Database & Theory & Theater & Motion Pictures & Politics & Restaurants \\
       \hline
       query & problem & theater & film & party & restaurant\\
       data & algorithm & play & movie & government & sauce\\
       database & bound & broadway & director & mayor & dish\\
        system & time & musical & hollywood & political & menu\\
       performance & polynomial & production & directed & election & food\\
       object & approximation & show & actor & president & dining\\
       \hline
    \end{tabular}
    \caption{Most discriminative features for some categories.}
    \label{tab:discriminative}
\end{table*}

\subsection{Word embedding}

As we presented in Section \ref{sec:novelty}, we introduce in a controlled manner one annotated category into our corpus at a rate defined by a logistic function. This way, it acts as our emerging new topic. All other categories evolve naturally in the corpus. While the frequency of a category increases with time, the frequency of each of the words of its lexical field also increases. We examine how vectors $v^w$ change when consecutive bunches of documents are used to update an embedding space. 

We built our embedding space using two vectorization techniques\footnote{Gold-Standard and Modelization techniques are available at \url{https://github.com/clechristophe/CEND}}: one built with Singular Value Decomposition (SVD) on Shifted PPMI matrices (SPPMI) and one built with Skip-Gram with Negative Sampling (SGNS) model. It has been demonstrated that SVD on SPPMI matrices yields results very close to Word2Vec approaches in terms of representation while insuring a certain stability \cite{antoniak2018evaluating}. Indeed, \cite{levy2014neural} showed that SGNS can be simplified into a matrix factorization problem using the SPPMI matrix defined as:
\begin{equation}\label{eq:sppmi}
     \hbox{SPPMI}(x,y) = \max\left\lbrace \log\frac{p(x,y)}{p(x)p(y)} - \log(s), 0\right\rbrace,
\end{equation}
where $s=15$ as is \cite{levy2014neural}.

In order to obtain stable word embeddings with SGNS at $t=0$ we initialized our space with a large subset ($T=12$ corresponding to a year of data in the NYTAC) at the beginning of our data. At each time step, the embedding space is updated with a bunch of documents keeping the same vocabulary than at initialization (no new words are introduced into the space) and the weights of the previous model is updated with new observations. For SVD, we do not need initialization and we start by building our SPPMI matrix on the first time step of our dataset. Meanwhile, for SVD, since we update the SPPMI matrix with new co-occurrences and then compute a new SVD, we need to add an alignment step in order to observe an interpretable movement in the embedding space. As in \cite{hamilton2016diachronic}, we use orthogonal Procrustes to align the learned embeddings of the new time-step with the previous model. 

\begin{figure*}
\begin{subfigure}{.5\textwidth}
  \centering
  \includegraphics[width=\linewidth]{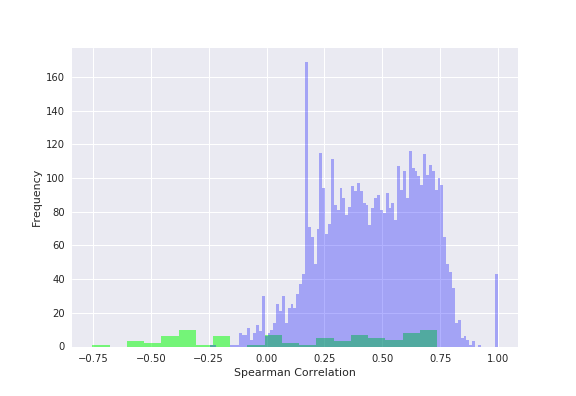}  
  \caption{SGNS}
\end{subfigure}
\begin{subfigure}{.5\textwidth}
  \centering
  \includegraphics[width=\linewidth]{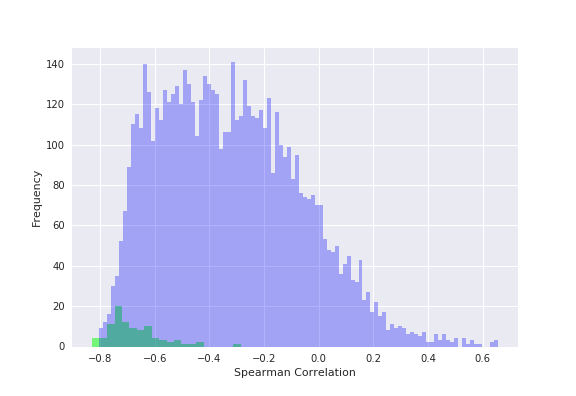} 
  \caption{SVD}
\end{subfigure}
\caption{Distribution of correlation between word frequencies and change magnitude over a subset of the vocabulary for two models. Distinction between emerging (green) and known words (blue) is highlighted.}
\label{fig:correlation}
\end{figure*}

\subsection{Key observation between word frequency and movement}

In this section, we show that correlation is a key component for detection emerging topics. First, we focus on the overall correlation between word frequency and word movement in the embedding space across the whole time span. To this end, we set $t = T$ and $n = T - 1$. In Figure \ref{fig:correlation}, we notice that, on one hand, a majority of words present a positive correlation for SGNS. On the other hand, they present a negative correlation for SVD. In other terms, in SVD, words sense tends to stabilize when their frequency increase while they are more volatile when using SGNS. We notice that some words present a strong and negative (close to -1) correlation and it seems that they are the ones related to the introduced category: their correlation distribution is highlighted in green in Figure \ref{fig:correlation}. Figure \ref{fig:correlation} present an example over one artificially introduced category but the observation is valid with all the categories in the NYTAC and SCI datasets.

\begin{figure*}
\begin{subfigure}{.5\textwidth}
  \centering
  \includegraphics[width=\linewidth]{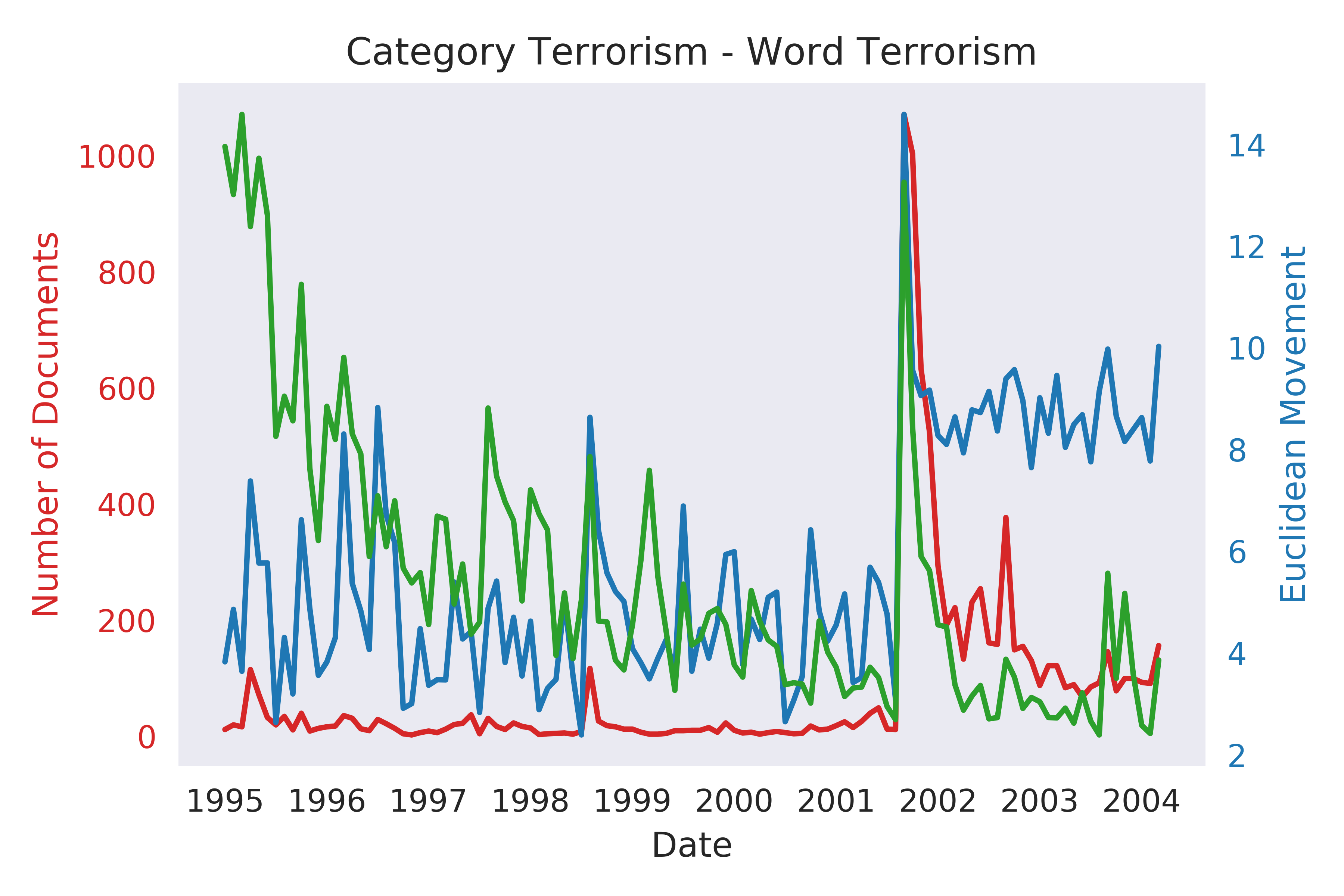}  
  \caption{Evolution of Terrorism category (red) and movement of word \textit{Terrorism} for SGNS (blue) and SVD (Green). The movement is \textbf{positively} correlated to the category frequency.}
\end{subfigure}
\hspace{2mm}
\begin{subfigure}{.5\textwidth}
  \centering
  \includegraphics[width=\linewidth]{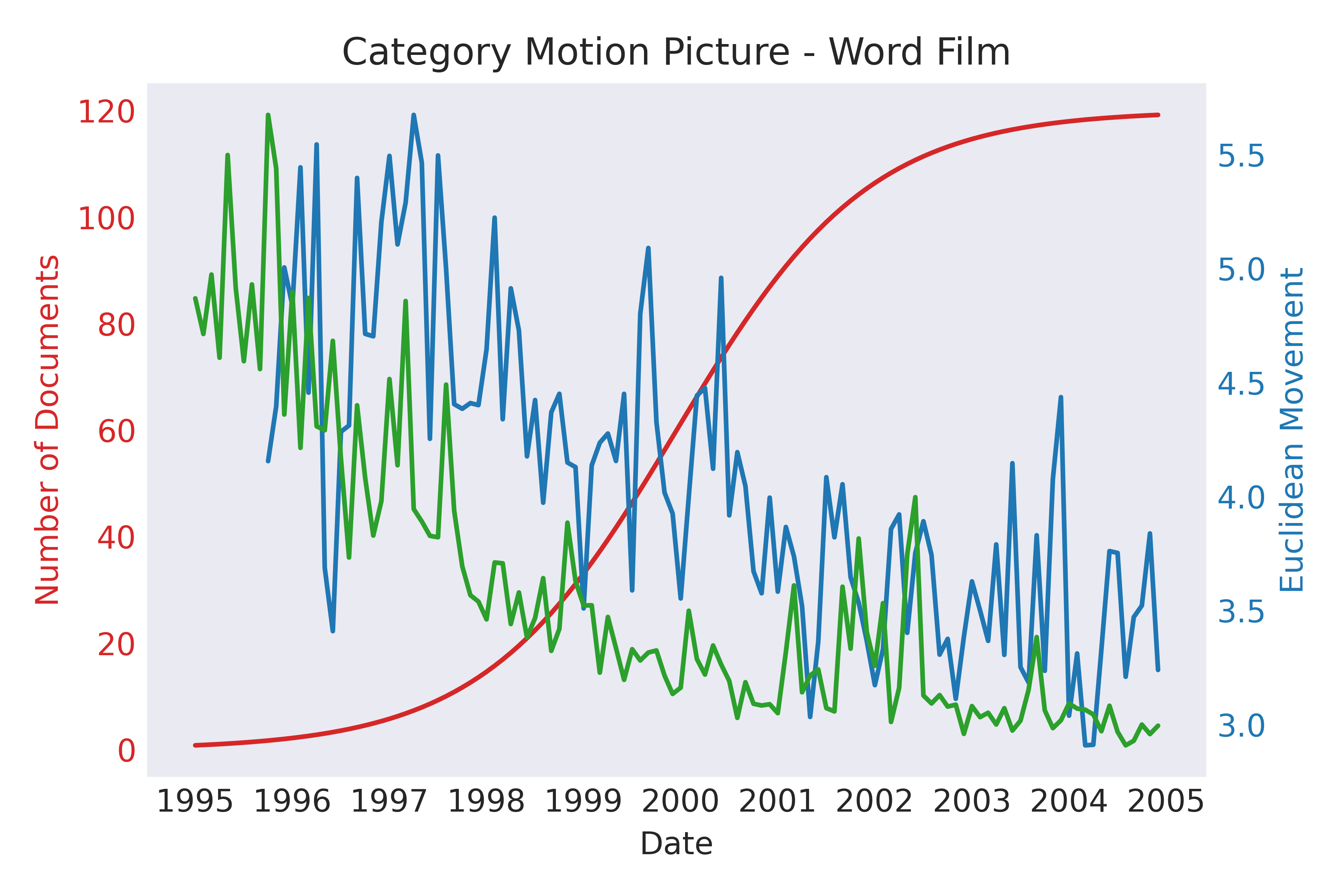}  
  \caption{Evolution of Motion Pictures category and movement of word \textit{Film} for SGNS (blue) and SVD (Green). The movement is \textbf{negatively} correlated to the category frequency}
\end{subfigure}
\caption{Difference between dynamics of an event (left) and an emerging topic (right).}
\label{fig:event}
\end{figure*}

Also, we notice that words related to events (appearing very fast) and to emergence (appearing slowly) move differently in the embedding space. In Figure \ref{fig:event}, we show the difference between the word \textit{Terrorism}, clearly linked with the \textit{Terrorism} category, which corresponds to an event-like dynamic around the 9-11 attacks, and the word \textit{Film} when we slowly introduced the \textit{Motion Pictures} category in our corpus. Word \textit{Terrorism}, which frequency increases suddenly after 9-11 attacks in the \textit{New York Times}, has an associated movement in the SGNS embedding space far superior at each time step than before the event. This word, without particular meaning change, is used in a very miscellaneous environment. This change of environment guides the word vector in the embedding space. In the SVD embedding space, movement is high during a brief moment around the event but comes back to a normal rate after the event. This observation allows us to assume that the modeling via SVD is more stable. The lower amplitude of the peaks for SVD supports this hypothesis that has already been studied in \cite{antoniak2018evaluating}. The word \textit{Film}, which frequency also increases but at a much slower rate, is associated with a decreasing movement through time. Even if the \textit{Motion Pictures} category exists at the beginning of our observation, this word is used in the same context through time and its meaning becomes more and more localized in the space. This observation, illustrated by these two words, seems to confirm our findings that words carried by event or emergence do not have the same behavior in the embedding space built with SVD or SGNS. 

\section{Results}\label{sec:results}

In this Section, we present the results obtained by our method for detecting words carried by a slow emerging topic. As we developed an artificially introduced signal (Section \ref{sec:novelty}) and a gold standard corresponding to this signal (Section \ref{sec:experiments}), it is possible to evaluate quantitatively with metrics such as Precision/Recall/F-Measure and AUC. Additionally to the general results, we investigate the effect of 3 parameters: the rate of emergence, the threshold and the size of the sliding window. Finally, we confirm, with the help of a control group, that our method effectively detects words linked to our artificially introduced topic.

\subsection{Baselines}

In order to evaluate our method CEND (as presented in equation \ref{eq:correlation}), we compare it to 4 baselines from the literature. Detecting weak signals associated with an emerging topic is a difficult task to evaluate quantitatively. We selected some baselines that work on a close problematic and are easily adaptable to work in our framework.

Our first baseline is adapted from \cite{allan2000detections} (TFIDF) where a method to detect and track new topics is presented. This method is based on the popular term frequency–inverse document frequency (TF-IDF) statistic and is built to raise alerts on particular terms when their TF-IDF statistics cross a manually-determined threshold. The second algorithm is \cite{xie2016topicsketch} (TopicSketch). It is an algorithm based on the monitoring of physical measurements (speed, acceleration) of textual entities (words and n-grams). It is built to raise alerts when these statistics have crossed a threshold. In \cite{huang2015topic} (HUPC), authors develop a method to extract representative patterns (e.g., words) of new emerging topics in microblog streams. After isolating patterns with a custom metric of utility, they determine if these patterns are from an emerging or known topic by comparing topics at each time-steps. \cite{peng2018emerging} (ET-EPM) uses the same metric of utility and combines it with a novelty measure based on the prediction of the evolution of a word. They use a graph analysis method to form emerging topics based on these isolated patterns. To describe their topics, they use hashtags available in their data. As we do not have hashtags in our datasets, we only used the extracted topic terms. 

\subsection{General Results}

In Table \ref{tab:results} we present the general results obtained by our method CEND with SGNS and SVD embeddings. We compare them with other baselines from the literature. For each method, we consider each word that was detected at least once during the entire observation period. While precision (P), recall (R) and F-measure (F) values are not particularly high, it is necessary to put them in context. When each category is introduced in our corpus, we try to detect its 100 most discriminative words in a vocabulary of size 20 000 for NYTAC and 5 000 for SCI. For each dataset, we evaluate our approach several times: we test each category as an emerging topic independently by introducing it as presented in Section \ref{sec:novelty}. Because we are shuffling documents for creating our emerging topics, we did the experiments 5 times by category and we present the mean results in Table \ref{tab:results}. For the two datasets, our method is the best in terms of F-measure while TopicSketch \cite{xie2016topicsketch} outperforms us in terms of precision in the NYTAC. The higher precision shown by TopicSketch can be explained by the trade-off between precision/recall. The method has a tendency to reduce detection errors by producing fewer alerts. However, its far lower recall shows that it misses a lot of the discriminative words. For SCI dataset, results are lower than for NYTAC partly because there is far less time step to analyze: as the corpus is separated into 15 years, our method has only 14 correlation scores for detecting the emergence of new words. Globally, results are quite similar for SGNS and SVD embeddings. While SVD embeddings are more stable, the use of correlation on smaller time-frame degrade global results on the entire observation.

\begin{table*}\centering
\begin{tabular}{|c|c|c|c||c|c|c|}
\hline
            & \multicolumn{3}{c||}{NYTAC} & \multicolumn{3}{c|}{SCI} \\ \hline
            & P      & R      & F      & P      & R      & F          \\ \hline
TFIDF \cite{allan2000detections}       &   0.17     &     0.12   &    0.14    &   0.10     &  0.12  &    0.11      \\ \hline
TopicSketch \cite{xie2016topicsketch} &   \textbf{0.48}     &  0.17 &  0.25      & 0.20  & 0.15   &  0.17       \\ \hline
HUPC \cite{huang2015topic} &  0.25 &   0.19   & 0.22    &     0.14   &    0.16    &    0.15        \\ \hline
ET-EPM \cite{peng2018emerging}  &    0.27    &  0.22      &  0.24   &   0.18     &  0.22      &   0.20     \\ 
        \hhline{|=|=|=|=|=|=|=|}
CEND-SGNS     &   0.37     &      0.33  &   0.34     &  0.22  &   0.32 & 0.26  \\ \hline
CEND-SVD     &   0.32     &      \textbf{0.45}  &  \textbf{0.37}     &  \textbf{0.24}  &   \textbf{0.36} & \textbf{0.29}\\ \hline
\end{tabular}
\caption{Mean average performance of novelty detection methods over 5 runs by dataset with a rate $r=0.5$.}
\label{tab:results}
\end{table*}

\subsection{Effect of the rate of emergence}

As we said in Section \ref{sec:novelty}, we introduce emerging topics into our datasets with a signal corresponding to a logistic function where we can control the rate $r$; We experimented with 3 values of $r$: $r=0.3$ corresponding to a slow rate of emergence, $r=0.5$ corresponding to a normal rate of emergence and $r=1$ corresponding to a fast rate of emergence closer to an event-like rate. As we see in Table \ref{tab:rate}, our models CEND-SGNS and CEND-SVD perform best with slow rates of emergence and our baselines are better when the rate of emergence becomes faster. This observation supports our hypothesis that the rate of emergence is a crucial parameter to take into account when choosing a type of approach: detecting slow emerging topics is not the same task as detecting events in a dataset.

\begin{table}[]
    \centering
    \begin{tabular}{c|c|c|c}
        \hline
         & $r=0.3$ & $r=0.5$ & $r=1.0$ \\ 
         \hline
        TFIDF & 0.11 & 0.14 & 0.18 \\
        TopicSketch & 0.19 & 0.25 & 0.28 \\
        HUPC & 0.14 & 0.22 & 0.24 \\
        ET-EPM & 0.16 & 0.24 & 0.27 \\
        \hhline{=|=|=|=}
        CEND-SGNS & 0.32 & 0.34 & 0.26 \\
        CEND-SVD & 0.36 & 0.37 & 0.27 \\
        \hline
    \end{tabular}
    \caption{Evolution of f-measure performance for different algorithms and different rates in the NYTAC dataset.}
    \label{tab:rate}
\end{table}

\subsection{Control Group}

In order to check if the detected words are detected because of their link with the artificially introduced emerging topic, we experimented with a control group where no emerging topic is expected. Instead of introducing a topic with the signal represented in Section \ref{sec:novelty}, we chronologically shuffled the category documents and introduced them in relation to a noisy signal. This way, we analyze the results in terms of Precision/Recall/F-measure in Table \ref{tab:control}.

\begin{table}
    \centering
    \begin{tabular}{cccc}
    \hline
       & P & R & F\\
    \hline
    NYTAC & 0.02 & 0.04 & 0.03 \\
    SCI & 0.02 & 0.02 & 0.02 \\
    \hline
    \end{tabular}
    \caption{Mean performance of CEND method under the control group}
    \label{tab:control}
\end{table}

The very low results seem to corroborate our initial hypothesis that the negative correlation we observed previously is linked with the emergence of a topic.

\subsection{Ranking Ability}

Our method CEND raises an alert each time the correlation between a word frequency and its movement in the embedding space is lower than a predefined threshold. These alerts have a monitoring purpose and their goal is to anticipate changes in the data. While we looked at general results for every word that was detected at least once, it is also interesting to study if some words have been detected \emph{several times}. By ranking words by their number of alerts associated with them, we can evaluate our model with the traditional information retrieval metrics: Receiver Operating Characteristic (ROC) Curve and Area Under Curve (AUC).

\begin{figure}[h]
   \includegraphics[width=\linewidth]{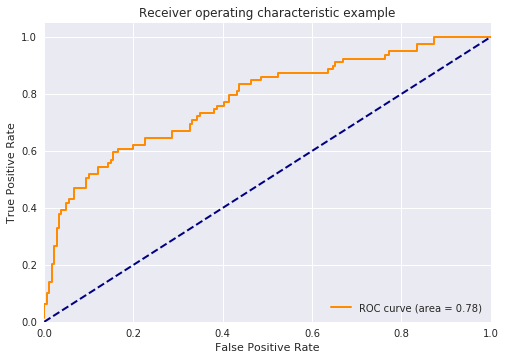}
    \caption{\label{fig:roc} ROC Curve when the category Motion Pictures is introduced in the NYT.}
\end{figure}

In Figure \ref{fig:roc}, we observe a ROC Curve for one category of the NYT and in Table \ref{tab:auc}, AUC for several categories of NYT and SCI as well as the most detected words are presented. Compared to the general results in Table \ref{tab:results}, we notice that our model is not very effective to detect all of the 100 words of our ground-truth but when it detects some of them, it has a tendency to detect several times so making them more visible. For SCI  dataset, general results are lower than for NYT but, for each category, we manage to keep an AUC between 0.70 and 0.85.

\begin{table}
\centering
    \begin{tabular}{cccc}
        \hline
       Category & AUC & $Word_{SNGS}$ & $Word_{SVD}$ \\
       \hline
           & & database & query\\
        Database & 0.71 & algorithm & data \\
          &  & access & database\\
        \hline
           & & general & problem\\
        Theory & 0.79 & constant & algorithm\\
            & & linear & polynomial\\
        \hline
            & & written & play\\
        Theater & 0.82 & character & broadway\\
            & & play & show\\
        \hline
    \end{tabular}
    \caption{Examples of AUC and most detected words for some categories in NYT and SCI.}
    \label{tab:auc}
\end{table}

\subsection{Parameter tuning}\label{sec:threshold}

Our method CEND depends on several parameters that are manually controlled: the threshold value $k$ and the size of the sliding window for computing the correlation $n$. For each corpus, we experimented and measured the optimal values for these parameters. For the threshold, the question is, once the correlation coefficient is computed, to determine whether it is significant or not. 
In order to generalize the threshold value to every category and every time-step, we compute variability intervals at each time-step. We do so by estimating the standard deviation $s$ thanks to the mean of the correlation over the vocabulary $\Bar{\rho}$. At each time-step, we compute a threshold value $k = -1.96 * s + \Bar{\rho}$ with $-1.96$ corresponding to the 97.5th quantile of a zero-centered gaussian. When testing with SVD embeddings on the NYT dataset, a decision rule evolving around $-0.65$ is obtained for all categories. For the size of the sliding window $n$, we tested with 4 values: $3,5,10,15$ and a sliding window of size 5 is best for NYTAC as we have enough time-step to compute it. For the SCI dataset, we used $n=3$ because we have 15 time-steps only.

\section{Discussion and Conclusion}\label{sec:conclusion}

In this work, we presented a method for detecting weak signals associated with slow emerging topics in textual streams. We designed simple experiments to test and measure quantitatively the performance of our approach. We analyzed the impact of hyper-parameters and showed that our method outperforms other algorithms from the literature. We based our method on the hypothesis that signals associated with slow emerging topics present a specific type of movement in embeddings spaces built with SVD and SGNS. We verified this hypothesis and observed that words associated with a slow emergence tend to present a negative correlation between their movement and frequency. We noticed that the positive correlation between movement and frequency represents a documented characteristic of word2vec: the more a word is used in a common context, the larger the norm of its vector is \cite{schakel2015measuring}. We also noticed that correlations are mostly negative when using SVD on the SPPMI matrix for modelization. This difference could be explained by the fact that SVD is a much more stable algorithm and less biased towards new observation as SGNS. This conclusion is supported in \cite{antoniak2018evaluating} but should deserve more analysis in our specific framework. In this work, we chose to use simple embeddings algorithms and not contextual approaches as BERT \cite{devlin2018bert} and ELMO \cite{peters2018deep} because we wanted to show that our observation is effective to detect novelty in a simple and inexpensive manner. Our observations (i.e. negative correlation) have been done on two different corpora and used for a detection task. It can be explained by the fact that a word is poorly defined when it has not been used enough in a corpus and its position in an embedding space become more precise when its frequency increases. Whether this observation is more largely valid or not should be confirmed by future research. In particular, we believe that the quality of the text may play an important role. In this work we only used curated documents with high quality writing coming from journalistic and scientific articles.

Several extensions for our method are easily conceivable. We showed in Figure \ref{fig:correlation} the differences in the distribution of correlation between novel and pre-existing words and it seems that these distribution could be separated by a statistical test inspired by \cite{blanchard2010semi}. Also, some approaches in the literature weight each word in relation to its part-of-speech tagging. We could imagine that it would make the detection task easier because discriminative words for a category tend to be nouns. It would be easier, for future industrial use, to automatically cluster detected words in order to better illustrate the detected emerging topic.

\bibliography{custom}
\bibliographystyle{acl_natbib}
\end{document}